# Development of a Testbed for Autonomous Vehicles: Integrating MPC Control with Monocular Camera Lane Detection

**Shantanu Rahman (190021101)**

**Nayeb Hasin (190021115)**

**Mainul Islam (190021312)**

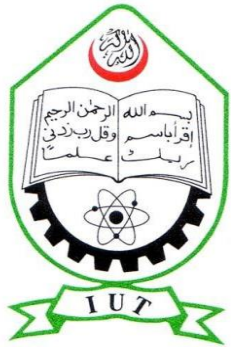

Department of Electrical and Electronic Engineering

**Islamic University of Technology (IUT)**

Gazipur, Bangladesh

June 2024

# Development of a Testbed for Autonomous Vehicles: Integrating MPC Control with Monocular Camera Lane Detection

Approved by

-------------------------------------------

**Professor Dr Golam Sarowar**

Supervisor and Professor,

Department of Electrical and Electronic Engineering,
Islamic University of Technology (IUT),
Boardbazar, Gazipur-1704.

Date:

# TABLE OF CONTENTS

# TABLE OF FIGURES

# LIST OF ACRONYMS

| | |
|---|---|
| LEMPC | Linear Error Model Predictive Control |
| LMPC | Linear Model Predictive Control |
| LQR | Linear Quadratic Regulator |
| LTV-MPC | Linear Time-Varying Model Predictive Control |
| MPC | Model Predictive Control |
| NEMPC | Nonlinear Error Model Predictive Control |
| NMPC | Nonlinear Model Predictive Control |
| PID | Proportional Integral Derivative |
| ROI | region of interest |
| ROS | Robot Operating system |
| ACC | Adaptive Cruise Control |
| ADAS | advanced driver-assistance systems |
| CAN | Controller Area Network |
| ECUs | Electronic Control Units |
| LASER | light wave |
| LiDAR | Light Detection and Ranging |
| RADAR | Radio Detection and Ranging |
| V2X | Vehicle-to-Everything |
| SPI | Serial Peripheral Interface |
| USART | Universal Synchronous/Asynchronous Receiver/Transmitter |
| CAN | Controller Area Network |
| ESC | Electronic Speed Controller |
| VGA | Video Graphics Array |
| MEMS | Micro-Electro-Mechanical Systems |
| SoC | System-on-Chip |

# ABSTRACT

Autonomous vehicles are becoming popular day by day not only for autonomous road traversal but also for industrial automation, farming and military. Most of the standard vehicles follow the Ackermann style steering mechanism. This has become to de facto standard for large and long faring vehicles. The local planner of an autonomous vehicle controls the low-level vehicle movement upon which the vehicle will perform its motor actuation. In our work, we focus on autonomous vehicles in road and perform experiments to analyze the effect of low-level controllers in the simulation and a real environment. To increase the precision and stability of trajectory tracking in autonomous cars, a novel method that combines lane identification with Model Predictive Control (MPC) is presented. The research focuses on camera-equipped autonomous vehicles and uses methods like edge recognition, sliding window-based straight-line identification for lane line extraction, and dynamic region of interest (ROI) extraction. Next, to follow the identified lane line, an MPC built on a bicycle vehicle dynamics model is created. A single-lane road simulation model is built using ROS Gazebo and tested in order to verify the controller's performance. The root mean square error between the optimal tracking trajectory and the target trajectory was reduced by 27.65% in the simulation results, demonstrating the high robustness and flexibility of the developed controller.

# Chapter 1

## RELATED WORKS

Model Predictive Control (MPC) has emerged as a crucial strategy in autonomous driving due to its capacity for handling complex multi-variable control problems and constraints. According to Falcone et al. (2007), MPC excels in trajectory planning by optimizing a vehicle's path over a finite time horizon, taking into account both vehicle dynamics and environmental constraints. This capability ensures that autonomous vehicles can plan effective trajectories that navigate safely and efficiently through dynamic environments. Moreover, as demonstrated by Camacho and Alba (2013), MPC's real-time adaptability allows it to adjust to changing conditions such as varying road surfaces and unexpected obstacles, maintaining safe and optimal vehicle operation.

The integration of MPC with vision-based systems enhances the autonomous vehicle's ability to perceive and react to its surroundings. Bender, Ziegler, and Stiller (2014) introduced the concept of lanelets, which can be integrated with MPC to improve lane-keeping performance using monocular cameras. These cameras are favored for their cost-effectiveness and ease of integration, processing images to identify lane markings and provide essential navigational information. Edge detection methods, such as those by Borkar et al. (2009), laid the groundwork for lane detection but struggled with lighting variability. Recent advancements, highlighted by Huval et al. (2015), leverage deep learning techniques, specifically convolutional neural networks (CNNs), to significantly enhance the accuracy and robustness of lane detection across diverse environmental conditions.

The synergy between MPC and advanced lane detection algorithms is pivotal for autonomous vehicle control. Studies by Kong et al. (2015) provide a comprehensive overview of kinematic and dynamic models used in MPC, outlining their advantages and limitations. For instance, hybrid lane detection systems that combine multiple cues, such as color and edge information (Lee & Lee, 2017), improve detection accuracy and robustness. Additionally, Zhang et al. (2017) explore MPC's application in cooperative driving scenarios, where multiple autonomous vehicles coordinate lane-keeping and obstacle avoidance. This body of research underscores the potential of combining MPC with sophisticated vision-based systems to achieve reliable and efficient autonomous driving solutions.

# Chapter 2

## INTRODUCTION

The idea of autonomous vehicles is transforming transportation by offering more accessibility, efficiency, and safety. But creating and implementing such vehicles comes with a lot of technical difficulties, mainly in the areas of control systems and perception of the surrounding environment. The goal of this research project is to enhance the performance and dependability of autonomous vehicle navigation by developing a testbed that combines monocular camera-based lane recognition with Model Predictive Control (MPC).

Robust control method Model Predictive Control (MPC) is well-known for its capacity to manage multivariable control issues with limitations. Using dynamic modeling, MPC forecasts a vehicle's future states and modifies control inputs to reach desired trajectories within operational and physical bounds. Because of its predictive power, MPC is especially well-suited to the complex and dynamic environments that autonomous vehicles must navigate.

A car's perception system relies heavily on lane detection to ensure correct placement on the road. With the use of advanced image processing techniques, monocular camera-based lane detection provides a cost-effective and adaptable solution by interpreting visual input from a single camera. Our testbed attempts to take advantage of the advantages of both technologies by combining MPC with monocular camera lane identification, leading to more reliable and adaptable vehicle control.

This project tackles a number of important issues in the development of autonomous cars, such as efficient algorithms for decision-making, real-time sensor data processing, and environmental perception. The testbed will act as a platform to address these problems, offering insightful information and laying the groundwork for future developments in autonomous driving technologies.

## 2.1 Background and Research motivation

In recent years, traffic accidents caused by driver error have increased rapidly, making autonomous driving technology a crucial solution (Shi and Sun, 2018). Many of these accidents result from vehicles crossing lane lines or losing control. In autonomous driving research, environmental perception and trajectory tracking technology are key to advancing the field. Consequently, many researchers have delved deeply into lane detection and trajectory tracking technologies (Bila et al., 2017).

In lane detection for structured roads, an improved RANSAC algorithm and double road pattern fitting method have been proposed. Simulation results demonstrate good real-time performance under various road conditions, but the study did not include region of interest (ROI) selection for images (Ozgunalp and Dahnoun, 2016). To improve lane detection accuracy, feature mapping based on IPM and parabolic fitting was introduced by Zheng et al. (2018), achieving a detection accuracy of 96%. However, this method's ROI extraction cannot adapt to changes in longitudinal vehicle speed. Ozgunalp et al. (2017) proposed an algorithm based on dense vanishing point estimation, significantly improving lane line detection accuracy, but it also lacks ROI extraction and adaptation to longitudinal speed changes. Lee and Kim (2016) suggested extracting lane lines using partial ROI to enhance efficiency in complex environments, yet this method also fails to adapt to variable speeds.

In trajectory tracking, various studies have utilized PID, LQR, and MPC algorithms for effective control (Guo et al., 2016). Li et al. (2016) proposed a PID steering controller and SMC drive controller, which perform well compared to traditional methods but are unsuitable for high-speed vehicles and cause shaking with speed changes. Levinson et al. (2011) applied an LQR controller to the integrated control of Stanford University's autonomous vehicles, but did not analyze the impact of vehicle dynamics constraints on optimal control. Attia et al. (2014) used nonlinear lateral and longitudinal dynamic coupling models for autonomous vehicle control, demonstrating good lateral controllability, but the method is computationally complex and lacks real-time performance. Cao et al. (2017) presented an optimal model-based trajectory tracking strategy to improve real-time performance, but it is effective only on paths with assumed curvature. Gutjahr et al. (2017) applied a linear time-varying model predictive controller for trajectory tracking optimization, but did not test the controller's performance or evaluate it in conjunction with real extracted lanes.

In summary, the mentioned lane detection research did not automatically extract ROI based on vehicle speed changes. Similarly, previous trajectory tracking studies did not use the extracted lane line as the expected trajectory, nor did they evaluate the combined effect of lane detection and trajectory tracking.

## 2.2 Objectives of our work

Our objective is to evaluate the performance of MPC under conditions of low computational power, low-resolution camera feedback, limited resources, and system noise, and compare it to previously popular methods to determine if there is any improvement. Our focus is on making the ROI extraction adapt to changes in vehicle speed and accurately track the extracted lane lines. First, we propose a method for dynamic ROI extraction of lane lines. Then, the extracted region is processed using sliding window line detection and quadratic curve fitting. Following this, we construct an MPC based on a bicycle model of vehicle dynamics to track the expected trajectory. Finally, we design a single-lane change scenario and an MPC controller with front wheel angle compensation to test control performance and enhance tracking accuracy.

## 2.3 Expected outcome of our research

Our work will bring out a cheaper testbed to test control system like MPC and to use in for research purposes. As the testbed will be cheaper, people will get easy access to it. Also, we aim to show the improvement to using MPC over PID as feedback loop using conducting experiments that depict real life scenario.

## 2.4 Methodology overview

Our methodology involves a series of tasks sequentially with a feedback system that predicts what should be the next optimal outputs that will interact with environment, for example the speed and steering of autonomous vehicle. Firstly, comes the input from the environment into the system. It primarily includes the image taken from the front of the autonomous vehicle. Simultaneously it takes speed and steering angle of the vehicle. Secondly the data are properly processed. For example, image taken goes through image processing, image calibration etc. Thirdly the processed data is fed into the NVIDIA JETSON NANO where MPC is run with the processed data, and corresponding output is sent to actuators that control the movement of the autonomous vehicle. Since this is a closed feedback loop, the mentioned series of tasks will happen infinitely like a loop until it reaches an objective, for example destination that lets the loop to be broken and the vehicle comes to a stop.

## 2.5. Autonomy: An overview of a modern car

Autonomous systems are hard to design. However, behind these almost magical systems there is articulated planning and robust communication which enables this feat to be possible. Before venturing further into the topic, a discussion of autonomy is a must.

### 2.5.1 Perception & Sensing

Like any other cyber systems which are autonomous, perception is the key. Acquiring data from the environment is the first obstacle we have to face before we decide on doing anything with it. For perception, myriads of sensors and methods are present in the modern car.

*Sonar*

Sonars are one of the oldest sensors for not only land vehicles but also other terrestrial and non-terrestrial vehicles. The sonars allow us to almost accurately predict the presence of an object nearby using ultrasonic waves in the medium. This is very popular in modern vehicles as a complementary to the looking glasses present already in the vehicle as part of the driver assist system.

*Light Detection and Ranging (LiDAR)*

LiDAR sensors are one of the most popular sensors in modern automobiles. LiDAR's working principle is very simple. It sends a light wave (LASER) towards a particular direction and senses it back, making sure of the occupancy around the sensors. LiDAR's are also capable of making decisions of distances based on the reflection caused by the object before them. LiDAR's unlike SONAR are very fast. Hence, the most popular LiDAR's rotate and sample this distance data around the vehicle making a simple 2D occupancy grid like information for the vehicle to make decisions on. Some LiDAR's are capable of taking these data in many planes, effectively measuring the whole environment around them like a 3D point cloud. This point cloud like representation is of great value to many autonomous vehicles' software.

*Radio Detection and Ranging (RADAR)*

Radio waves are one of the most fundamental gifts of modern engineering. Lights are also part of radio waves, in theory LiDAR's are a family of RADAR. So, why is RADAR more prevalent in modern autonomous systems? RADARs can be tuned in different frequencies, it's imperative that different frequencies react differently to different materials. This allows us to do a soft segmentation on the data during the collection of data itself. Also, most RADAR work in very high frequencies. This allows us to have much better point cloud information close by without interfering with other vehicles navigation system.

### 2.5.2 Actuation & Vehicles

Actuation in vehicles is integral to the operation and performance of various systems, converting energy into mechanical motion to control different components. Electrical actuators, such as solenoids and electric motors, are widely used in fuel injectors, starter motors, power steering, and window regulators. Hydraulic actuators, including hydraulic cylinders and pumps, are essential for braking systems like ABS and power steering. Pneumatic actuators find their applications in air braking systems and air suspension, particularly in heavy-duty vehicles. Each type of actuator offers specific advantages suited to different vehicle functions, ensuring precise control and enhanced safety.

- Electrical actuators: solenoids, electric motors
- Hydraulic actuators: hydraulic cylinders, pumps
- Pneumatic actuators: air cylinders, motors

- Applications: fuel injectors, starter motors, power steering, window regulators, ABS, air braking, suspension

*Advanced Actuation Technologies*

Advanced actuation technologies are revolutionizing modern vehicles, providing greater control, efficiency, and functionality. Drive-by-wire systems replace traditional mechanical controls with electronic systems, offering improvements in throttle, braking, and steering. Adaptive Cruise Control (ACC) uses actuators to maintain speed and safe following distances, while autonomous driving relies heavily on actuators for steering, acceleration, and braking. Active aerodynamics adjust spoilers and vents to enhance stability and efficiency, and active suspension systems improve ride quality by dynamically adjusting damping forces. These advancements ensure vehicles are more efficient, responsive, and capable of supporting cutting-edge features.

- Drive-by-wire: throttle-by-wire, brake-by-wire, steer-by-wire
- Adaptive Cruise Control (ACC)
- Autonomous driving: steering, acceleration, braking
- Active aerodynamics: adjustable spoilers, air vents
- Active suspension: dynamic damping forces

### 2.5.3 Communication systems in a modern vehicle

Communication systems in modern vehicles are complex networks that enable various components and systems to interact seamlessly, ensuring safety, efficiency, and driver convenience. At the heart of these systems is the Controller Area Network (CAN) bus, a robust, real-time communication protocol that connects various Electronic Control Units (ECUs) such as the engine control module, transmission control module, and infotainment system. Beyond the CAN bus, vehicles incorporate other networks like Flex Ray, Ethernet, and LIN (Local Interconnect Network) to handle different tasks ranging from high-speed data transfer for advanced driver-assistance systems (ADAS) to simpler, low-speed communications for comfort features like seat adjustments and lighting. These networks facilitate critical functions such as vehicle diagnostics, real-time data sharing for collision avoidance, and multimedia streaming. Additionally, modern vehicles are increasingly equipped with wireless communication technologies, including Bluetooth, Wi-Fi, and cellular networks, enabling features like hands-free calling, internet access, over-the-air software updates, and remote diagnostics. Integration with external systems via V2X (Vehicle-to-Everything) communication enhances situational awareness, allowing vehicles to exchange information with other vehicles, infrastructure, and networks, thereby improving traffic management, safety, and driving experience. This intricate web of communication systems underscores the sophistication and interconnected nature of today's automotive technology.

### 2.5.4 Stages of planning (Global & Local Planners)

In the context of autonomous vehicles, planning is a crucial aspect that ensures safe, efficient, and reliable navigation. This process typically involves various stages, with global and local planners playing distinct but complementary roles. Here's a detailed look at these stages:

*Global Planning*

Global planning is the overarching process that sets the high-level path and goals for the vehicle. It focuses on long-term decisions and navigation strategies over a broader scope. The global planner determines the overall route from the starting point to the destination. It might also optimize a circuit like path, in case of something like a race car. It may also take into account the terrain, roads, obstacles which are static and find a viable and efficient path. In most vehicle, global planning is done via road information and GPS coordinates. The vehicle will try to follow this generated trajectory in the long run. This is more of a high level than real-time trajectory planning. Thus, a global planner gives the general sense of direction when calculating the overall route.

*Local Planning*

Local planning focuses on real-time, short-term decisions that ensure the vehicle can safely and smoothly follow the global path. It deals with immediate obstacles and dynamically changing conditions. Local planning is always done in real time and is the low level trajectory controller actually responsible for steering and actuating the UGV. It is imperative the local planner is robust and may dynamically adjust to the vehicles current state and the environment and navigate accordingly. Local planner as expected is the main motion controller of the vehicle. In our work, we focus on the local planner and how to improve this local planner to suite the need of an Ackermann UGV.

*Interaction between Global and Local Planners*

The interaction between global and local planners is essential for seamless vehicle operation. The global planner provides a high-level path that the local planner follows. The local planner's role is to ensure that the vehicle adheres to this path while handling immediate obstacles and constraints. Local planning may also feed information back to the global planner, allowing for dynamic adjustments to the overall route based on real-time conditions. For example, if the local planner detects a roadblock, it can request a new route from the global planner. The global planner aims for overall efficiency and route optimization, the local planner ensures safety and smooth navigation by handling immediate and detailed decision-making.

2.5.6 Stability & Robustness

*Stability*

**Stability** refers to the ability of a control system to return to its equilibrium state after a disturbance. A stable system will not exhibit unbounded or oscillatory behavior over time.

- **Types of Stability**:
    - **Absolute Stability**: A system is absolutely stable if all its poles (roots of the characteristic equation) have negative real parts. This implies that all solutions to the system's differential equations decay to zero over time.
    - **Relative Stability**: A system is relatively stable if it remains close to its equilibrium point, despite small disturbances. It is often assessed by the damping ratio and natural frequency of the system's response.
- **Criteria for Stability**:

- **Lyapunov Stability**: A system is stable if there exists a Lyapunov function $V(x)$ such that $V(x)$ is positive definite and its time derivative $\dot{V}(x)$ is negative definite. This criterion is useful for nonlinear systems. (Raković & Levine, 2019)
- **Routh-Hurwitz Criterion**: A mathematical method used to determine the stability of a linear time-invariant (LTI) system by analyzing the sign changes in the characteristic polynomial's coefficients [14].
- **Nyquist Criterion**: A graphical method to assess the stability of a feedback system based on the open-loop frequency response. It involves plotting the Nyquist diagram and checking for encirclements of the critical point (-1,0) in the complex plane [15].

*Robustness*

**Robustness** refers to a system's ability to maintain performance despite uncertainties, variations, and disturbances. A robust system is designed to function correctly under a range of operating conditions and parameter variations.

- **Sources of Uncertainty**:
    - **Model Uncertainty**: Differences between the actual system model and the assumed model, including parameter variations.
    - **Disturbances**: External influences, such as noise, external forces, or environmental changes.
- **Robust Control Strategies**:
    - **H-infinity (H∞) Control**: A framework that seeks to minimize the worst-case gain of the closed-loop system, providing a balance between performance and robustness against disturbances and uncertainties. [16]
    - **Madden's Criterion**: A method used to design controllers that ensure robustness by modifying the system dynamics to handle uncertainties effectively. [17]
    - **Sliding Mode Control**: A robust control technique that forces the system's state to evolve along a predefined sliding surface, making the system insensitive to parameter variations and external disturbances. [18]

2.5.7 Closing the loop: Control Theory

To properly actuate the vehicle we need to "close the loop". In the realm of control theory, "closing the loop" refers to the concept of feedback control, where the output of a system is fed back into the system as input to achieve desired performance. This feedback mechanism is essential for maintaining stability, improving accuracy, and enhancing the overall performance of control systems. Two prominent control strategies that exemplify this concept are Proportional-Integral-Derivative (PID) control and Model Predictive Control (MPC).

*PID Control*

PID control is one of the most widely used control strategies due to its simplicity and effectiveness. It is characterized by three parameters: Proportional (P), Integral (I), and Derivative (D), each contributing to the control effort in a specific way:

- Proportional (P): This term produces an output value that is proportional to the current error value. It helps to reduce the overall error but can lead to steady-state error if not tuned correctly.
- Integral (I): This term accounts for the accumulation of past errors. It integrates the error over time, helping to eliminate any residual steady-state error. However, too much integral action can cause instability and oscillations.
- Derivative (D): This term predicts future error based on its rate of change, providing a damping effect that helps to reduce overshoot and oscillations. It improves the system's response speed and stability.

PID control is almost always easy to implement and can be tuned intuitively. PID control is used practically in the industry and is a de facto standard for most machineries and robots in the field for its high versatility as a controller. Depending on the system, however, if the system behaves non-linearly, PID parameters may become hard to optimize.

*Model Predictive Control (MPC)*

MPC is a more advanced control strategy that uses a model of the system to predict future behavior and optimize control actions over a finite horizon. Unlike PID control, which relies on past and present error, MPC considers the system's dynamic model and future constraints to determine the optimal control input. MPC uses a mathematical model of the system to predict future states based on current inputs and disturbances. At each time step, MPC algorithm solves an optimization problem to find the control inputs that minimize a cost function, typically involving the deviation from desired control state and effort. MPC, thus, naturally handles constraints on the inputs, states, and outputs, ensuring the control actions stay within safe and feasible bounds.

MPC offers superior performance by anticipating future behavior and adjusting control actions proactively. Depending on the variant of the algorithm, MPC is capable of handling complex, nonlinear systems and multivariable control problems. However, MPC is a resource hungry control algorithm and if the mathematical model of MPC is not that accurate, the controller may perform worse.

*Comparing PID and MPC*

- **Complexity**: PID control is straightforward and easier to implement, while MPC is more complex and requires a good model of the system.
- **Performance**: MPC generally provides better performance, especially for systems with constraints and those requiring precise control over a horizon. PID control is adequate for simpler systems with minimal constraints.
- **Tuning and Implementation**: PID controllers are easier to tune and implement, whereas MPC may require sophisticated algorithms and computational resources for real-time optimization.

## 2.6. Camera Calibration

Cameras are the main sensory input in our system. However, like all other sensors, cameras are also imperfect. Thus, to perfectly get clear and accurate information from the camera, we need to fix the camera images independent of the camera parameters. This can be done via assuming a linear pinhole projection model for the camera. To mathematically model a pinhole camera, we make the following assumptions:

1. The image plane, which is the surface that captures the light rays, is positioned in front of the pinhole. In reality, it is located behind the pinhole, but this assumption simplifies the projection model by eliminating the need to invert the image.

2. All light rays from various points converge at the pinhole, also referred to as the center of projection or the camera center.

3. The concept is that the image of a point is the projection of that point onto the image plane, determined by where the line from the camera center to the point intersects the image plane.

The world space encompasses the camera, points, and objects.

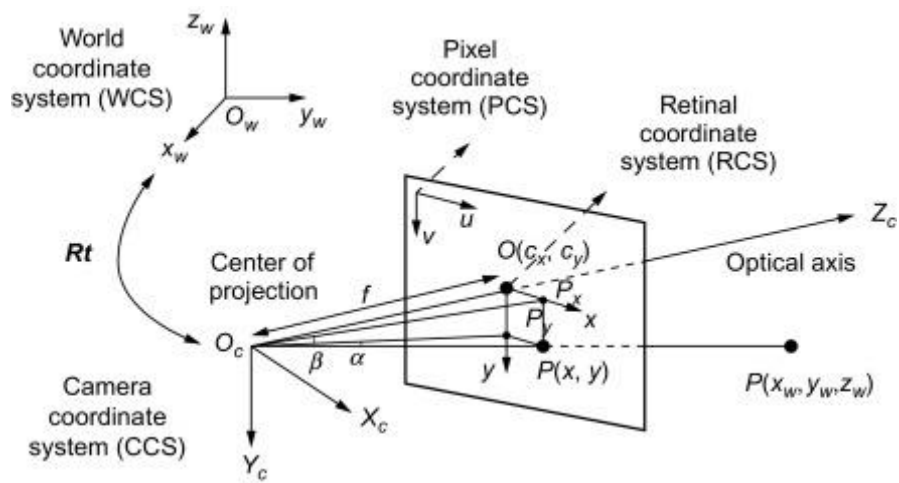

**Figure 1: Pinhole camera model**

The figure above illustrates a basic configuration of the pinhole camera model. In this setup, the camera center is positioned at the origin, and the image plane is parallel to the XY plane, situated at a specific distance from the origin (this distance is known as the focal length, which will be explained later).

2.6.1 Extrinsic Calibration

Extrinsic calibration determines the transformation between the world coordinate system and the camera coordinate system. It describes the camera's position and orientation in the world. This transformation is composed of:

1. **Rotation Matrix (R)**: This matrix represents the orientation of the camera relative to the world coordinate system. It describes how the world coordinates need to be rotated to align with the camera's view.
2. **Translation Vector (T)**: This vector represents the position of the camera in the world coordinate system. It describes how the origin of the world coordinates needs to be translated to coincide with the camera's origin.

Together, the rotation matrix and translation vector form a transformation matrix that converts world coordinates to camera coordinates. This transformation allows us to view the world from the camera's perspective.

2.6.2 Intrinsic Calibration

Intrinsic calibration determines the transformation from the camera coordinate system to the image plane of the camera. It describes the camera's internal parameters, which affect how it projects 3D points onto the 2D image plane. The intrinsic parameters include:

1. **Focal Length**: This defines the distance between the camera's lens and the image plane. It influences the scale of the projected image.
2. **Principal Point**: This represents the coordinates of the point where the optical axis intersects the image plane. It is usually close to the center of the image.
3. **Skew Coefficient**: This parameter accounts for any non-rectangular pixels in the image sensor. It is often zero for most modern cameras.
4. **Distortion Coefficients**: These parameters account for lens distortion, including radial and tangential distortion, which affect the accuracy of the projection.

The intrinsic parameters are typically represented in a camera matrix, which, combined with distortion coefficients, defines how 3D points in the camera coordinate system are projected onto the 2D image plane.

2.6.3 Camera Calibration Process

The process of camera calibration involves determining both the extrinsic and intrinsic parameters:

1. Take multiple images of a known calibration pattern (e.g., a checkerboard) from different angles and positions.
2. Identify and accurately detect the feature points on the calibration pattern in each image. In our case, the feature points were crosses on a chessboard.
3. Use the known positions of the feature points in the world and their corresponding positions in the images to solve for the extrinsic and intrinsic parameters. This is typically done using optimization algorithms that minimize the projection error.

**ROS camera Calibration**:

Calibration involves using the interior vertex points of a checkerboard, where a "9x7" board, for example, employs an interior vertex parameter of "8x6." This process should be conducted in a well-lit, unobstructed area measuring 5m x 5m that contains checkerboard patterns. The calibration is performed using a monocular camera that publishes images through the ROS framework.

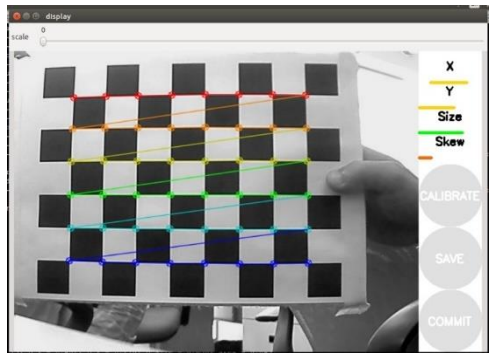

Figure 2: Intrinsic Camera Calibration

To achieve accurate calibration, we needed to move the checkerboard around within the camera frame to ensure it appeared on the camera's left, right, top, and bottom fields of view. This involved positioning the checkerboard horizontally (X bar) and vertically (Y bar) across the field of view, adjusting its distance and tilt relative to the camera (Side bar), and ensuring it filled the entire field of view. Additionally, we tilted the checkerboard left, right, top, and bottom to account for skew. At each position, we held the checkerboard steady until the image was properly highlighted in the calibration window.

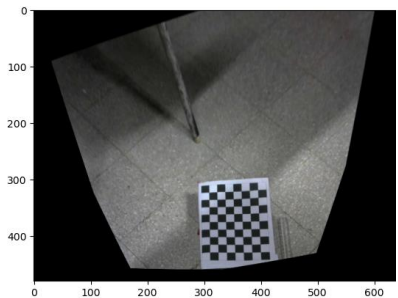

Figure 3: Extrinsic Camera Calibration

A successful calibration we got results in real-world straight edges appearing straight in the corrected image.

## 2.7 Lane detection

The feature-based method utilizes edges and local visual traits, such as gradients, color, brightness, texture, orientation, and variations, which remain fairly stable despite road shape variations but are sensitive to changes in lighting. The process of lane detection and tracking using images and sensors depends on the sensors installed on the vehicle and the camera feed. Initially, image frames undergo pre-processing, and a lane detection algorithm is used to identify lane markings, while sensor data aids in determining the path. This technique employs inverse projective mapping (IPM) to generate a bird's-eye view of the road, with a Hough transform for lane detection and a sliding window method for lane tracking. The road image is first converted to grayscale and blurred temporally. Applying IPM provides a bird's-eye perspective, where lanes are identified by detecting pairs of parallel lines separated by a certain distance. These IPM images are then binarized, following homography and perspective warp on the saturation channel of the HSV panel, and divided into two halves. A one-dimensional matched filter of the histogram is applied to each sample to find the line center, selecting pixels with high correlation exceeding a threshold as lane centers. The sliding window method then tracks the lane, concatenating indices and extracting lane line pixel positions.

### 2.7.1 Reading and Resizing the Calibration Image
Reads and resizes images to a standard size (640x480 pixels) for consistency in subsequent processing steps

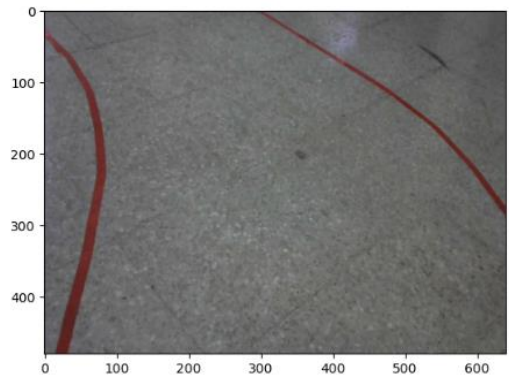

**Figure 4: Reading and Resizing**

### 2.7.2 Camera Calibration and Distortion
Camera calibration involves defining the camera matrix and distortion coefficients, which are essential for correcting lens distortions. The undistorted image is then obtained using these parameters.

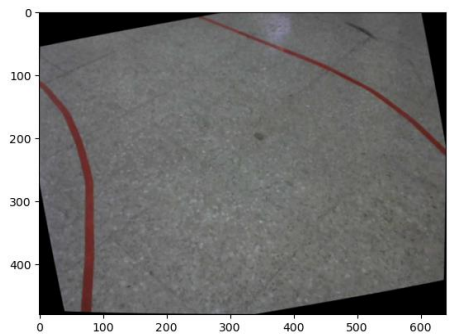

**Figure 5: Camera Calibration and Distortion**

### 2.7.3 Reading, Resizing, Rotating, and Undistorting the Ground Image
Rotates the ground image by 10 degrees and undistorts it to correct for lens distortion, making it ready for further processing.

### 2.7.4 Defining Source and Destination Points for Perspective Transformation

Defines the source and destination points to map a region of interest from the original image to a new perspective, simulating a change in the viewpoint.

### 2.7.5 Finding the Homographic Matrix and Applying Perspective Transformation

Computes the homographic matrix that maps the source points to the destination points. This matrix is then used to warp the image, transforming it to the new perspective.

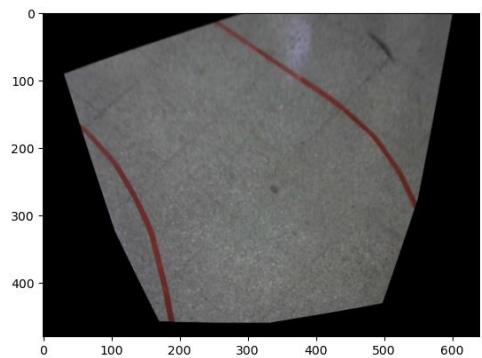

**Figure 6: Perspective Transformation**

### 2.7.6 Convert Image to HSV and Split Channels:

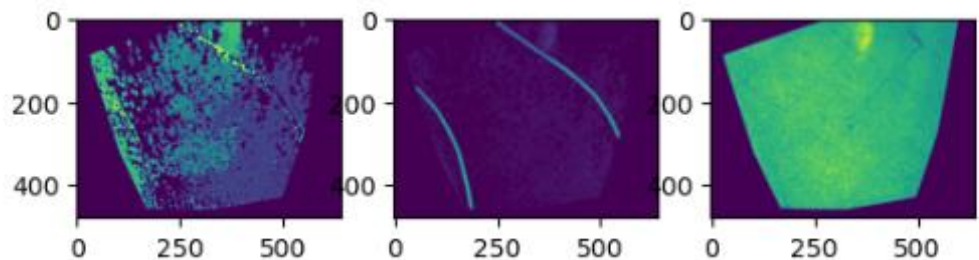

**Figure 7: HSV and Split Channels**

☐ Convert the warped image from BGR to HSV color space.

☐ Split the HSV image into its three channels: Hue (H), Saturation (S), and Value (V).

☐ Display each channel separately.

2.7.7 Blur and Threshold the Saturation Channel:

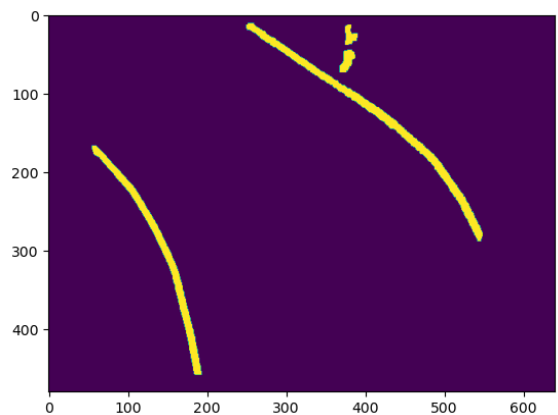

**Figure 8: Blur and Threshold the Saturation Channel**

- ☐ Applies a Gaussian blur to the Saturation channel to reduce noise.
- Thresholds the blurred image to create a binary image, where pixels with a value greater than 30 are set to 255 (white) and others to 0 (black).
- Displays the thresholded image.

2.7.8 Morphological Operations:

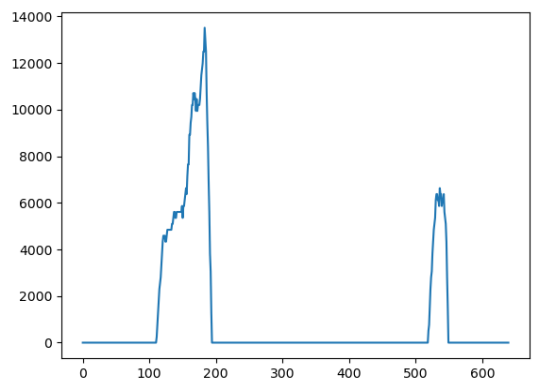

**Figure 9: Morphological Operations**

- Applies a morphological opening operation to the thresholder image to remove small noise and close gaps.
- Displays the morphologically processed image.

## 2.7.9 Sliding Window Algorithm for Lane Line Detection:

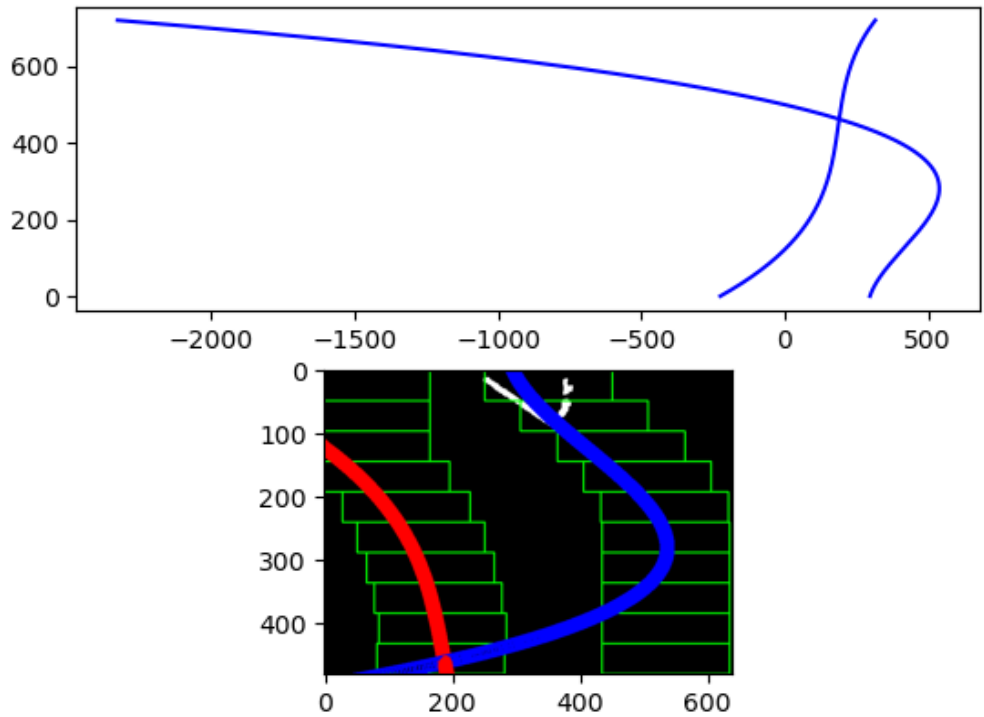

Figure 10: Sliding Window Algorithm

- Calculates and plots the histogram of the lower half of the morphologically processed image.
- Identifies the base positions for the left and right lanes using the histogram peaks.

## 2.7.10 Lane Line Polynomial Fitting
- Concatenates the identified pixel indices for both lanes.
- Fits a 3rd-degree polynomial to the identified lane pixels for both left and right lanes.

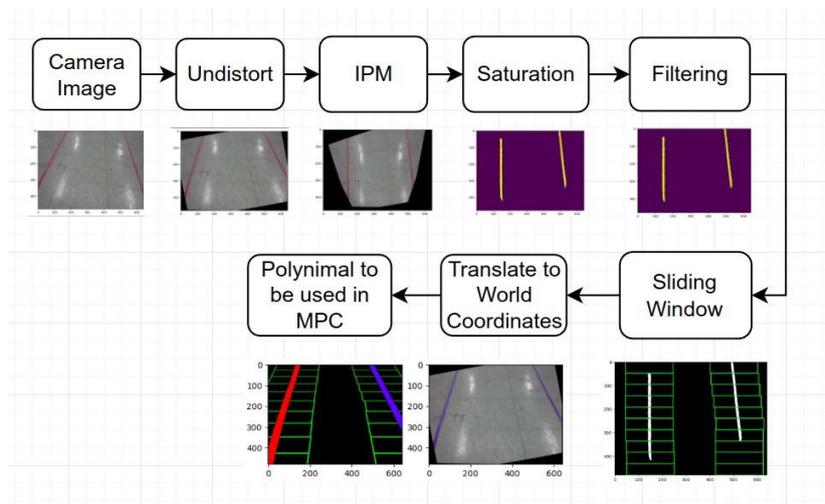

Figure 11: Overall lane detection pipeline

## 2.8 Controlling the vehicle

The execution competency of an autonomous system, often referred to as motion control, is crucial for translating intentions into actions. Its main goal is to implement planned intentions by supplying the necessary inputs to the hardware, which then produces the intended movements. Controllers interpret interactions with the real world in terms of forces and energy, while cognitive navigation and planning algorithms concentrate on the vehicle's speed and position relative to its surroundings. Measurements within the control system are essential for evaluating performance, enabling the controller to respond to disturbances and adjust system dynamics to achieve the desired state. Utilizing system models can provide detailed descriptions of the desired motion, which is vital for effective motion execution.

### 2.8.1 Classical Control

Feedback control is the most common controller structure used in many applications. It relies on measuring the system's response and actively compensates for deviations from the desired behavior. This approach helps mitigate the negative effects of parameter changes, modeling errors, and unwanted disturbances. Feedback control can also influence a system's transient behavior and address measurement noise. [19], [20]

The most prevalent form of classical feedback control is the Proportional-Integral-Derivative (PID) controller, widely used in the process control industry. The PID controller's concept is straightforward and doesn't require a system model; instead, it operates based on the error signal using the control law:

$$u(t) = k_d \dot{e} + k_p e + k_i \int e(t)\, dt$$

Where, $e$ is the error signal, and $k_p$ $k_i$ and $k_d$ are the proportional, integral, and derivative gains of the controller, respectively [21]-[23].

However, feedback-only controllers have several limitations. The primary limitation is their delayed response to errors, as they only react once errors occur. Additionally, feedback controllers have a coupled response, meaning they address disturbances, modeling errors, and measurement noise through the same mechanism. It is more effective to handle the response to a reference separately from the response to errors [24].

By incorporating a feedforward term into the controller, another degree of freedom is introduced. This controller architecture, shown in Figure 5, helps overcome feedback control limitations. The feedforward term is added to the control signal without relying on any measurement of the controlled system, although it may involve measuring disturbances. Designing a feedforward control requires a comprehensive understanding of the physical system, often utilizing a model reference. Combining feedforward and feedback terms results in a two-degree-of-freedom controller [25]-[28].

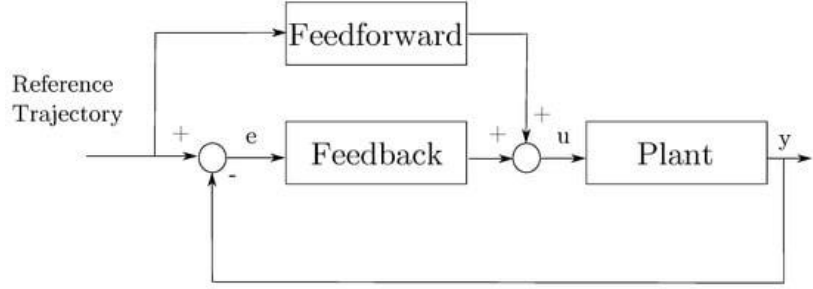

**Figure 11: 2 DOF Controller.**

State space control, often called modern control, is a method that aims to control the entire system vector by examining its states as a whole. This field is extensive, with ongoing research continuously expanding its boundaries. A linear state space model is typically expressed as:

$$\dot{x}(t) = A(t)x(t) + B(t)u(t)$$

$$y(t) = C(t)x(t) + D(t)u(t)$$

where $x(t)$ represents the system state vector, $u(t)$ is the control input vector, and $y(t)$ is the system output.

In autonomous systems, observations are generally nonlinear, necessitating the linearization of the state space equations for analysis [29]:

$$\dot{x}(t) = f(x(t), u(t))$$

$$y(t) = h(x(t), u(t))$$

A two-degree-of-freedom controller can be applied to nonlinear systems. Here, feedforward control generates a reference trajectory, while feedback control compensates for disturbances and errors [30]. To linearize the nonlinear system around a reference trajectory $x_r(t)$, we derive the linearized error dynamics:

$$\delta\dot{x}(t) = A(t)\delta x(t) + B(t)\delta u(t)$$

$$\delta y(t) = C(t)\delta x(t) + D(t)\delta u(t)$$

In these equations, $A, B, C, D$ are the relevant Jacobians [31]. If a trajectory generation process can create a reference input $u_r(t)$ that produces a feasible trajectory satisfying the nonlinear system dynamics, state space controllers can be designed to provide feedback compensation for the linearized error dynamics.

### 2.8.2 Model Predictive Control

Autonomous systems require motion models for planning and prediction. These models can also be utilized in control execution. A control method that leverages system modeling to optimize performance over a future time period is known as Model Predictive Control (MPC). The basic structure of MPC is illustrated in Figure 6. MPC is designed to combine the performance benefits of optimal control with the robustness of robust control. Typically, predictions are made over a short period called the prediction horizon, during which the model predictive controller aims to find the optimal solution. The model, and consequently the controller, can be adjusted in real-time to adapt to changing conditions.

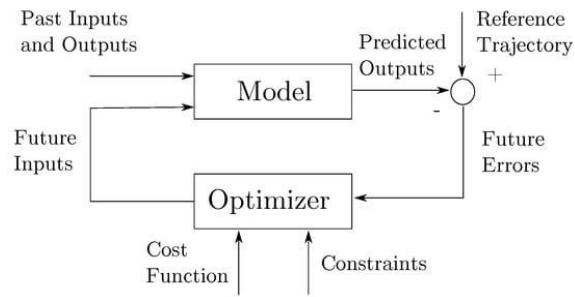

**Figure 12: MPC Controller.**

Model predictive control (MPC) has achieved significant success in industrial process control applications, primarily due to its straightforward concept and its capability to handle complex process models with input constraints and nonlinearities effectively. MPC offers several attractive features, including the ability to design multivariable feedback controllers easily and specify system inputs, states, and outputs that the controller must regulate. Furthermore, MPC allows for defining an objective function to optimize control efforts, managing time delays, rejecting disturbances both measured and unmeasured, and leveraging stored information on expected future conditions. These capabilities are particularly beneficial for repetitive tasks such as maintaining a consistent path, mimicking natural processes by integrating optimization and feedback adjustments.

In automotive applications, MPC has also seen widespread adoption to improve fuel efficiency, reduce emissions, and enhance overall vehicle safety. However, implementing MPC in automotive systems presents unique challenges compared to the industrial process control sector. Automotive systems operate with much shorter sampling periods, typically in milliseconds, and face constraints on computing resources due to space limitations. Thus, advancements in processor speed, memory capacity, and the development of innovative algorithms are crucial for expanding the use of MPC in the automotive industry.

The general formulation of the model predictive control (MPC) problem is defined as follows:

Objective:

Minimize the cost function $J(U(t)) = F(x(N \mid t)) + \sum_{k=0}^{N-1} L(x(k \mid t), y(k \mid t), u(k \mid t))$

Subject to:

1. System dynamics: $x(k+1 \mid t) = f(x(k \mid t), u(k \mid t))$

2. Output equation: $y(k \mid t) = h(x(k \mid t), u(k \mid t))$

3. State constraints $x_{\min} \leq x(k \mid t) \leq x_{\max}$, for $k = 1, \ldots, N_c$

4. Output constraints $y_{\min} \leq y(k \mid t) \leq y_{\max}$ for $k = 1, \ldots, N_c$

5. Control input constraints: $u_{\min} \leq u(k \mid t) \leq u_{\max}\backslash)$, for $k = 1, \ldots, N_c u$

6. Initial state: $x(0 \mid t) = x(t)$

7. Future control inputs: $u(k \mid t) = \kappa(x(k \mid t))$ for $k = N_u, \ldots, N-1$

Here, $t$ is the discrete time index. The vector notation $v(h \mid t)$ represents the value of $v$ predicted at $h$ time steps ahead, based on information up to time [32]. Equations (7) and (8) describe the discrete-time dynamics of the system with a sampling period $T_s$, where $x \in \mathbb{R}^n$ is the state, $u \in \mathbb{R}^m$ is the control input, and $y \in \mathbb{R}^p$ is the output [33].

The control input sequence $U(t) = (u(0 \mid t), \ldots, u(N-1 \mid t))$ is the optimizer, where $N$ is the prediction horizon. The cost function, similar to optimal control, includes the stage cost $L$ and terminal cost $F$ State and output constraints are enforced over the horizons $N_c$ and $N_c u$ respectively. The control horizon $N_u$ represents the number of optimized steps before applying the terminal control law [34].

In each control cycle at time $t$ the MPC strategy proceeds as follows: system outputs are measured, and the state $x(t)$ is estimated. This state estimate initializes the optimization problem in Equation (6) and enforces the initial condition in Equation (12) [35]. After solving the MPC optimization problem, the optimal input sequence $U^*(t)$ is obtained, and the first element $u^*(0 \mid t)$ is applied to the system, i.e., $u(t) = u^*(0 \mid t)$ In the next cycle, the process repeats using the updated state estimate, thereby applying feedback control [36].

## 2.9 Vehicle Dynamics

2.9.1 Ackermann Steering Geometry

Ackermann steering geometry is a design principle used in the steering system of vehicles to ensure that all wheels correctly follow the intended circular path during a turn. The purpose is to reduce tire wear and increase the vehicle's stability by making sure that the inner and outer wheels turn at appropriate angles [37].

*Key Concepts*

- **Steering Angles:** The inner and outer wheels need to turn at different angles to follow the arcs of circles with different radii [38].
- **Turning Radius:** The path followed by the wheels forms arcs of circles with different radii, with the inner wheel following a smaller radius and the outer wheel a larger one.
- **Geometric Relationship:** Ackermann geometry establishes a geometric relationship between the steering angles of the front wheels.

Mathematical Equation

For a vehicle with a wheelbase $L$ and a track width $W$ [39]:

Let $\theta_{in}$ and $\theta_{out}$ be the steering angles of the inner and outer wheels respectively, and R be the turning radius of the vehicle's path. The Ackermann principle ensures that:

$$\cot(\theta_{in}) - \cot(\theta_{out}) = \frac{W}{L}$$

Where:

- $\theta_{in}$ = steering angle of the inner wheel.
- $\theta_{out}$ = steering angle of the outer wheel.
- $L$ = distance between front and rear axles (wheelbase).
- $W$ = distance between the left and right wheels (track width).

Derivation

1. The inner wheel follows a circle of radius $R_{in}$ and the outer wheel follows a circle of radius $R_{out}$
2. These radii are related by the track width $W$: Rout= $R_{out} = R_{in} + W$
3. For the vehicle to turn without slipping, the wheels should follow these paths:

$$R_{in} = \frac{L}{\tan(\theta_{in})}$$

$$R_{out} = \frac{L}{\tan(\theta_{out})}$$

4. Substituting $R_{out} = R_{in} + W$:

$$\frac{L}{\tan(\theta_{out})} = \frac{L}{\tan(\theta_{in})} + W$$

5. Simplifying: $\frac{1}{\tan(\theta_{out})} = \frac{1}{\tan(\theta_{in})} + \frac{W}{L}$
6. Using the cotangent identity:

$$\cot(\theta_{out}) = \cot(\theta_{in}) + \frac{W}{L}$$

### 2.9.2 Bicycle Model Vehicle Dynamics

The bicycle model simplifies the vehicle's dynamics by representing the vehicle with a single front wheel and a single rear wheel. It is widely used for analyzing and designing control systems for vehicles due to its simplicity and effectiveness in capturing essential dynamics.

Key Assumptions

- The vehicle is approximated by a two-wheel model, with one front and one rear wheel.
- The vehicle's mass is distributed uniformly.
- The model assumes small angles to linearize the equations of motion.

Mathematical Equations

The bicycle model focuses on the lateral (side-to-side) and yaw (rotational) dynamics of the vehicle.

State Variables

- $\beta$: Sideslip angle (the angle between the vehicle's heading and its velocity vector).
- $\psi$: Yaw angle (rotation about the vertical axis).
- $v$: Velocity of the vehicle.
- $a_f\ a_r$ : Lateral accelerations at the front and rear wheels respectively.
- $\delta$: Steering angle of the front wheel.

Equations of Motion

The bicycle model consists of two primary equations: one for lateral motion and one for yaw motion.

1. **Lateral Motion:**

$$m(\dot{v}_y + v_x\dot{\psi}) = F_{yf} + F_{yr}$$

Where:

- $m$: Mass of the vehicle.
- $v_y$ : Lateral velocity.
- $v_x$ : Longitudinal velocity.

- $F_{yf}$ : Lateral force on the front tire.
- $F_{yr}$ : Lateral force on the rear tire.

2. **Yaw Motion:**

$$I_z \ddot{\psi} = aF_{yf} - bF_{yr}$$

Where:

- $I_z$ : Yaw moment of inertia.
- $\ddot{\psi}$ : Yaw acceleration.
- $a$: Distance from the center of gravity to the front axle.
- $b$: Distance from the center of gravity to the rear axle.

Lateral Tire Forces

The lateral tire forces $F_{yf}$ and $F_{yr}$ are modeled as linear functions of the slip angles $\alpha_f$ and $\alpha_r$ :

$$F_{yf} = C_f \alpha_f$$

$$F_{yr} = C_r \alpha_r$$

Where:

- $C_f$ : Cornering stiffness of the front tire.
- $C_r$ : Cornering stiffness of the rear tire.

Slip Angles

The slip angles for the front and rear tires are given by:

$$\alpha_f = \delta - \frac{v_y + a\dot{\psi}}{v_x}$$

$$\alpha_r = -\frac{v_y - b\dot{\psi}}{v_x}$$

# Chapter 3

## HARDWARE SETUP

In the initial setup, we used an STM32 microcontroller to coordinate control functions with an Arduino, interfaced with a brushed ESC. Visual feedback was provided by an OV7670 camera, while an MPU-9250 IMU handled motion sensing. An ESP32 module managed wireless communication. In version 2, we upgraded to an Nvidia Jetson Nano as the central processing unit, with a Pixhawk managing actuation control and IMU measurements. A buck converter reduced the 12V LiPo battery voltage to 5V for the Jetson Nano. Communication between the Jetson Nano and Pixhawk utilized MAV Link messages. Lane detection was achieved using a USB webcam connected to the Jetson Nano, and brushed ESCs were used to control motor speeds and steering.

## 3.1 Version 1

**Version 1 Overview:**

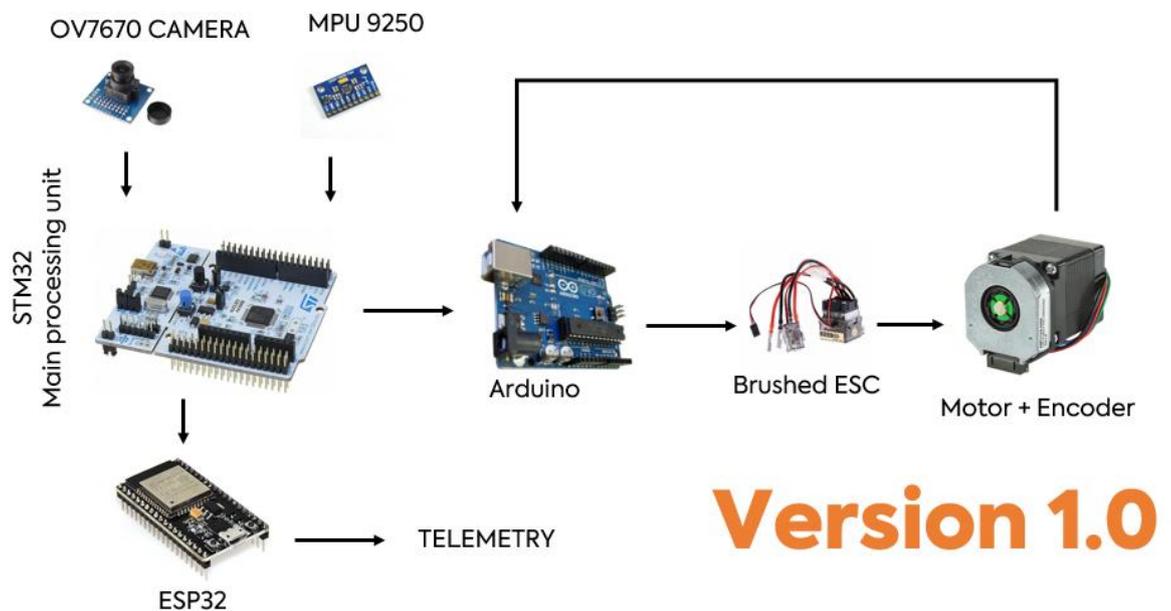

Figure 13: Version 1 Architecture

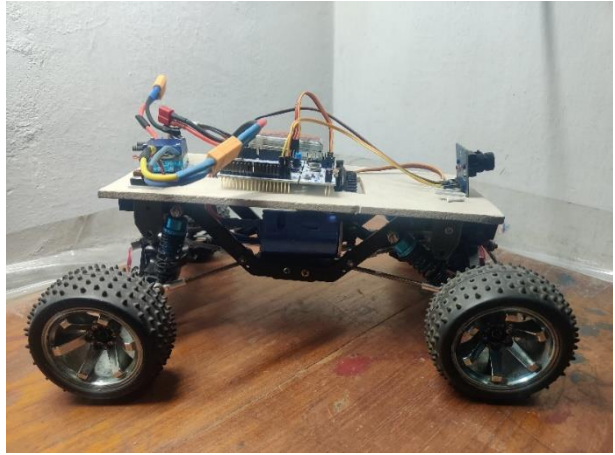

**Figure 14: Version 1 hardware**

The detailed description of the components is given bellow:

### 3.1.1 STM32 Microcontroller

- **Description**: The STM32 is a family of 32-bit microcontrollers based on the ARM Cortex-M processor. It is known for its high performance, low power consumption, and rich set of peripherals.
- **Applications**: Used as the central control unit for handling complex control algorithms and interfacing with other components.
- **Features**:
    - High processing power with clock speeds up to 400 MHz.
    - Various communication interfaces, including SPI, I2C, USART, and CAN.
    - Low power modes for energy-efficient operation.

### 3.1.2 Arduino Board

- **Description**: Arduino is an open-source electronics platform based on simple software and hardware. It is widely used for prototyping and building custom electronics projects.
- **Applications**: Acts as the intermediary controller between the STM32 and the brushed ESC, managing motor control logic.
- **Features**:
    - Easy-to-use development environment with a vast library support.
    - Digital and analog I/O pins for connecting various sensors and actuators.
    - Compatibility with a wide range of shields and modules for extended functionality.

### 3.1.3 Brushed ESC (Electronic Speed Controller)

- **Description**: An ESC is a device that controls the speed of a DC motor, typically used in RC cars, drones, and other robotics applications.
- **Applications**: Controls the speed and direction of the DC motor used for propulsion.
- **Features**:
    - PWM (Pulse Width Modulation) input for speed control.

- Supports a range of voltages and current ratings suitable for different motor specifications.
- Typically includes over-current protection and thermal shutdown features.

### 3.1.4 OV7670 Camera

- **Description**: The OV7670 is a low-cost, low-resolution camera module with an integrated image sensor. It is widely used for basic image capture in embedded systems.
- **Applications**: Provides visual feedback to the system, enabling image processing and object recognition.
- **Features**:
    - VGA resolution (640x480 pixels) with a frame rate of up to 30 fps.
    - Supports a variety of image formats and compression modes.
    - Uses a parallel interface for communication with microcontrollers.

### 3.1.5 MPU-9250 IMU (Inertial Measurement Unit)

- **Description**: The MPU-9250 is a 9-axis MEMS sensor combining a 3-axis accelerometer, a 3-axis gyroscope, and a 3-axis magnetometer.
- **Applications**: Provides motion tracking and orientation data, crucial for stabilizing and controlling the system.
- **Features**:
    - Wide range of measurement capabilities, including acceleration, angular velocity, and magnetic field sensing.
    - High accuracy and low noise performance.
    - Digital communication interfaces, including I2C and SPI.

### 3.1.6 ESP32 Module

- **Description**: The ESP32 is a low-cost, low-power system-on-chip (SoC) with integrated Wi-Fi and Bluetooth capabilities. It is highly popular for IoT and wireless communication projects.
- **Applications**: Handles wireless communication and telemetry, enabling data exchange between the system and a remote control or monitoring station.
- **Features**:
    - Dual-core processor with clock speeds up to 240 MHz.
    - Integrated Wi-Fi (802.11 b/g/n) and Bluetooth (Classic and BLE).
    - Rich set of peripherals, including ADCs, DACs, SPI, I2C, UART, and GPIOs.

## 3.2 Version 2

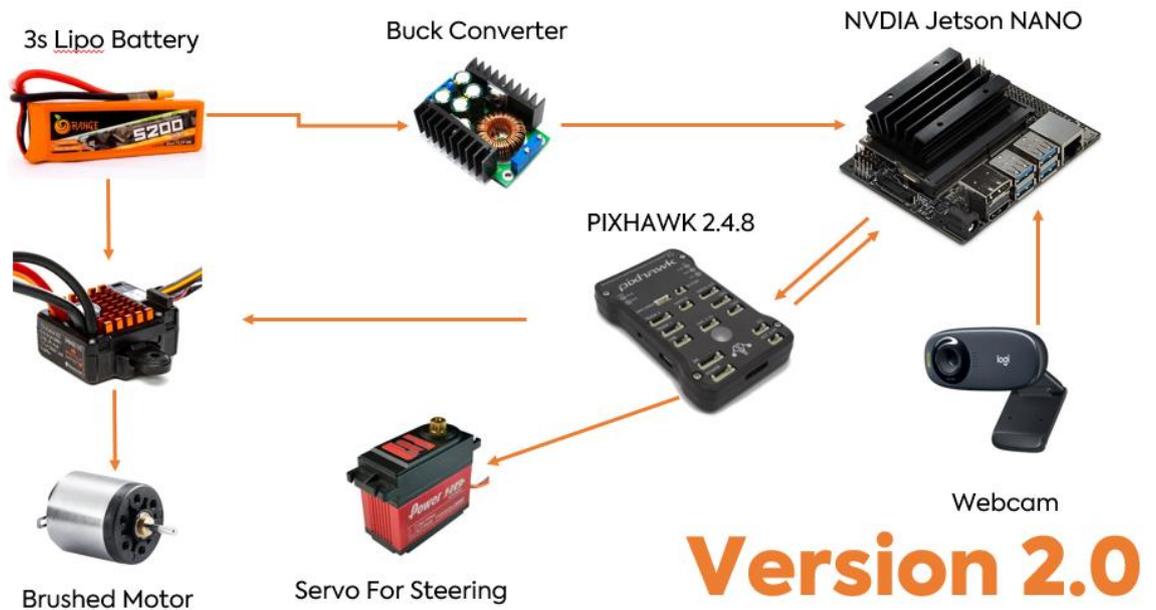

Figure 15: Version 2 Architecture

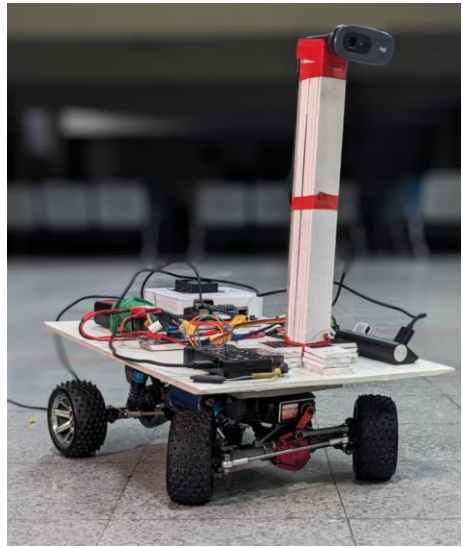

Figure 16: Version 2 hardware

### 3.2.1 Nvidia Jetson Nano

- **Description**: The Nvidia Jetson Nano is a small, powerful computer designed for AI applications. It features a GPU, CPU, and RAM, making it capable of running deep learning models and handling high-performance computing tasks.
- **Applications**: Serves as the central processing unit, handling image processing, computer vision, and AI algorithms for the system.
- **Features**:
    - Quad-core ARM Cortex-A57 CPU and 128-core Maxwell GPU.

- o 4 GB LPDDR4 RAM.
- o Support for Ubuntu Linux, CUDA, and cuDNN for deep learning applications.
- o Connectivity options include HDMI, USB, Ethernet, and GPIO pins.

### 3.2.2 Pixhawk

- **Description**: The Pixhawk is a popular open-source flight control hardware platform for drones and robotic systems. It supports a wide range of sensors and has robust connectivity for various peripherals.
- **Applications**: Used for actuation control and IMU measurements, providing stable flight control and sensor integration.
- **Features**:
    - o 32-bit microcontroller with multiple sensor inputs and outputs.
    - o Supports various communication protocols, including I2C, SPI, and PWM.
    - o Compatible with a wide range of autopilot software, such as PX4 and ArduPilot.
    - o Integrated IMU for real-time orientation and motion sensing.

### 3.2.3 Buck Converter

- **Description**: A buck converter is a type of DC-DC converter that steps down voltage from a higher level to a lower level efficiently.
- **Applications**: Converts the 12V LiPo battery voltage to 5V, supplying power to the Jetson Nano.
- **Features**:
    - o High efficiency, typically above 90%.
    - o Adjustable output voltage with current regulation.
    - o Compact size suitable for integration into small spaces.

### 3.2.4 MAVLink Communication

- **Description**: MAVLink (Micro Air Vehicle Link) is a lightweight communication protocol for drones and unmanned vehicles. It facilitates communication between the flight controller and ground control stations or other components.
- **Applications**: Enables seamless communication between the Jetson Nano and Pixhawk, allowing data exchange and control commands.
- **Features**:
    - o Simple and efficient message format for sending commands and telemetry data.
    - o Supports various message types, including GPS data, sensor readings, and control commands.
    - o Widely adopted in the UAV community, ensuring compatibility with many systems and software.

### 3.2.5 USB Webcam

- **Description**: A standard USB webcam used for video capture and image processing. It is connected to the Jetson Nano to provide visual input.

- **Applications**: Used for lane detection and image processing tasks, providing real-time video feed to the Jetson Nano.
- **Features**:
  - High-definition video resolution, typically up to 1080p.
  - USB connectivity, easy to interface with the Jetson Nano.
  - Built-in microphone and support for various video streaming protocols.

### 3.2.6 Brushed ESC (Electronic Speed Controller)

- **Description**: An ESC is a device that regulates the speed of a DC motor by varying the voltage supplied to it. Brushed ESCs are commonly used in RC vehicles and drones.
- **Applications**: Controls the motor speeds and powers the servo for steering control.
- **Features**:
  - PWM input for precise speed control.
  - Supports a range of voltage and current ratings suitable for different motors.
  - Includes safety features such as over-current protection and thermal shutdown.

### 3.2.7 Power Servo

- **Description**: A power servo is an actuator that uses a motor and gearbox to provide precise rotational control. It is commonly used for steering in robotics and RC vehicles.
- **Applications**: Powers the steering mechanism, allowing precise control of the vehicle's direction.
- **Features**:
  - High torque output suitable for steering applications.
  - Fine control over angular position, typically adjustable via PWM signals.
  - Built-in feedback mechanism for position control and stability.

## 3.2 Inertial Measurement Unit:

The Pixhawk is a versatile flight control system that integrates an Inertial Measurement Unit (IMU), which includes accelerometers and gyroscopes. To get started, ensure the Pixhawk is properly connected and powered. Use ground control software such as Mission Planner or QGroundControl to configure the Pixhawk. Within the software, you should calibrate the IMU by following the calibration procedures for the accelerometer and gyroscope, ensuring accurate data collection. This calibration is essential for precise orientation and stability control of the vehicle. Additionally, confirm that the firmware, such as PX4 or ArduPilot, is installed and configured correctly to handle sensor data.

Reading IMU Data Using MAVLink

To access IMU data, the Pixhawk communicates using the MAVLink protocol. You can write scripts to read MAVLink messages that contain IMU data, such as RAW_IMU and ATTITUDE.

## 3.3 Inter system communication

3.3.1 MAVLink Communication

1. **Overview of MAVLink Protocol**:

    - **Definition**: MAVLink (Micro Air Vehicle Link) is a lightweight, header-only messaging protocol designed for communication between drones, ground control stations (GCS), and other unmanned vehicles.
    - **Purpose**: It provides a standardized way to send commands, telemetry data, and status updates, making it easy to integrate different components and software systems.

2. **Message Types**:

    - **Command Messages**: Used to send control commands from the ground station or central processing unit to the Pixhawk, such as SET_POSITION_TARGET_LOCAL_NED for position control or ARM to arm the motors.
    - **Telemetry Messages**: These messages include sensor data, status updates, and diagnostic information, such as ATTITUDE, RAW_IMU, and GPS_RAW_INT.

3.3.2 PWM Signal Generation

1. **Role of Pixhawk in PWM Control**:

    - **PWM Signals**: Pulse Width Modulation (PWM) signals are used to control the speed of brushless motors and the position of servos. These signals are typically 50 Hz pulses with varying duty cycles.
    - **Control Channels**: The Pixhawk outputs PWM signals to control up to 8 channels for motors and servos. Each channel corresponds to a specific control function, such as motor throttle, elevator, aileron, and rudder.

2. **PWM Signal Output**:

    o **Configuration**: Set up the Pixhawk's firmware (PX4 or ArduPilot) to define the mapping of PWM channels to specific outputs. This is usually done through the ground control software or configuration files.
    o **Example Configuration**:
    - **PX4 Configuration**: Use QGroundControl or modify the param file to set the PWM output parameters.
    - **ArduPilot Configuration**: Use Mission Planner or the APM Planner to adjust the MOTOR_PWM_MIN, MOTOR_PWM_MAX, and other related parameters.

### 3.3.3 Integration Flow

1. **Sending Commands from the CPU to Pixhawk:**

    - The central processing unit (e.g., Nvidia Jetson Nano) sends control commands via MAVLink messages to the Pixhawk. These commands can include navigation waypoints, control modes, and sensor data requests.

**Pixhawk Processing and Output**:

- Upon receiving the MAVLink messages, the Pixhawk processes the commands and converts them into PWM signals. These signals are then sent to the respective motor and servo control pins.

# Chapter 4

## SIMULATION TESTING

Before actually testing on the testbed, it is always a good idea to test the effects of the controller in a controlled environment where every information can be perceived and analyzed. That is why a software-based physics simulation was done before actually doing the hardware implementation. For the simulation, we used Gazebo in Ubuntu 20.04 (Coire i7, NVidia 2040).

## 4.1 Overview of technologies

In our case we used a combination of ROS and Gazebo for the simulation. But this wasn't the only available option. There are many options for physics-based simulation in the wild. MATLAB is the first choice for many researchers. MATLAB allows us to simulate vehicle dynamics and Simulink as well as show it in its UE4 plugin window (Islam & Okasha, 2019). However, MATLAB is slow and the licensing for the toolboxes are difficult to attain. CARLA Simulator is another simulation where physics-based simulation is implemented. CARLA is written with python with some extensive list of API and Customizability (Zhou et al., 2023), (Morais & Aguiar, 2022). However, CARLA's support for external model and real-life physics accuracy is subpar. NVidia's Isacc Sim is also a great tool. This is especially good for reinforcement learning as Isacc Sim is capable of simulating multiple robots of the same category at once (Ackerman, 2022). In our case, it is not necessary as we are not using RL. There are also other simulators available. Game engines, like unity3d or UE4, may be also used, but this is not standard practice in robotics. The standard approach in these cases is the integration of ROS tightly coupled with Gazebo Simulation Ecosystem.

## 4.2 Robot Operating System

The Robot Operating System is a versatile framework for building robotic systems. Developed initially in 2007 by Willow Garage, it has since become an open-source platform maintained by the Open Robotics organization. What sets ROS apart is its flexibility and modularity, designed to facilitate the development of complex robot behaviors across various robotic platforms. By using ROS, we can keep the components separate and modular. This keeps the simulation logic; the controller logic and lane detection logic separate. In this case, all the controller logic was written in C++ as a ROS Node, later ported to Jetson Nano.

## 4.3 Gazebo Simulation

The Gazebo simulation was implemented in another standard ROS package. The body was designed as an URDF. Initially, the joints were made using a PID controller to simulate the suspension. Later, it was removed from the system, as no noticeable change was observed during it.

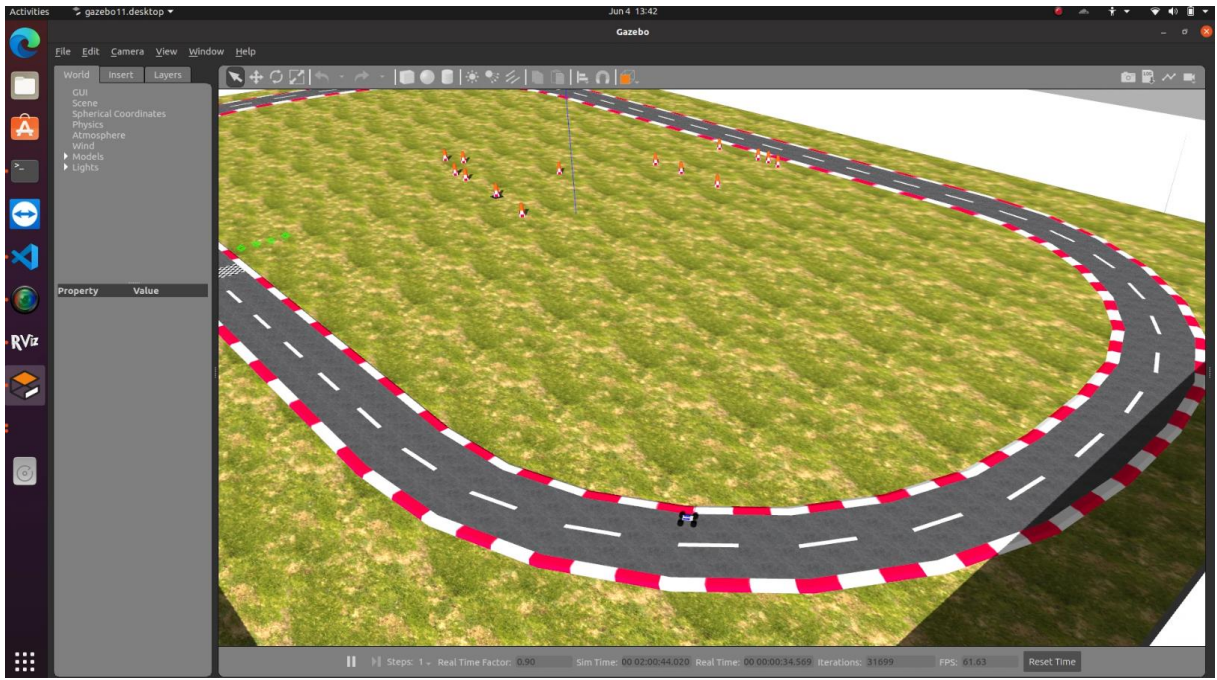

**Figure 17: Gazebo Environment**

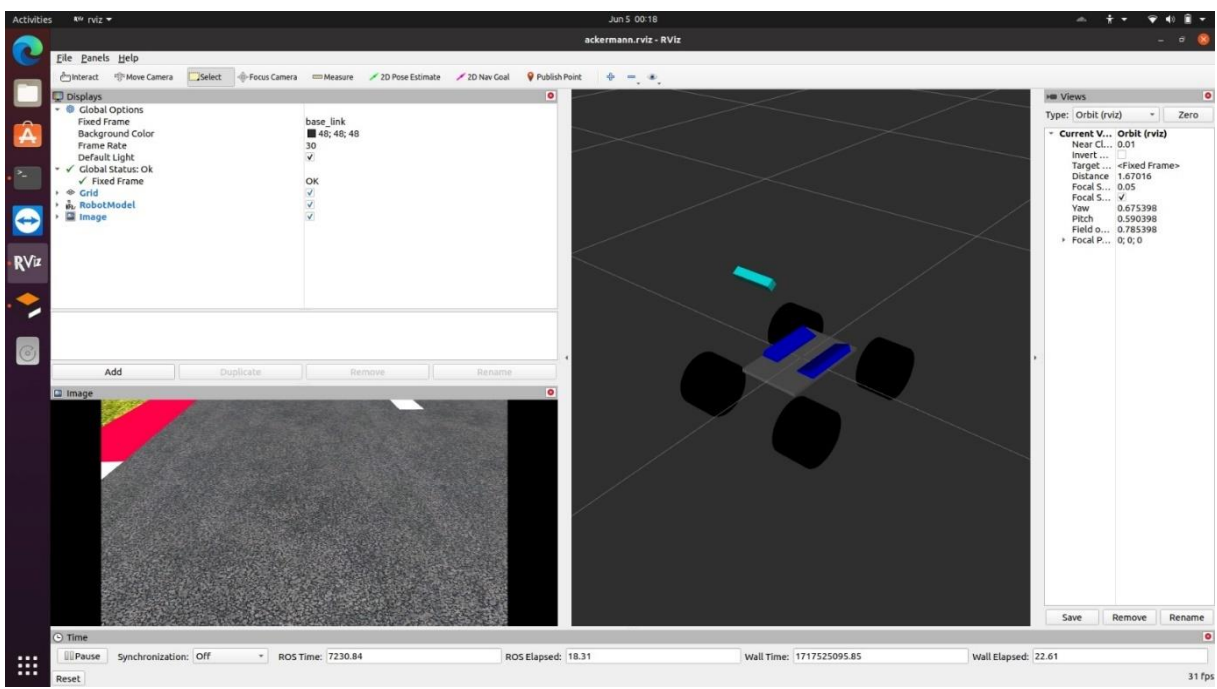

**Figure 18: Rviz Simulation**

## 4.4 ROS Architecture

The simulation was done using the AWS Robemaker environment in gazebo. The car model URDF was designed to realize a 1:10 model vehicle. The whole world was rendered in Gazebo using ROS. The URDF was published in a robot describer node. A gazebo Ackermann joint controller plugin was used to move the car and steer it. The joint controller PID was tuned to reflect the hardware setup. The gazebo controller subscribed to the /Ackermann topic. The lane detection and MPC controller node published to this topic making the vehicle steerable.

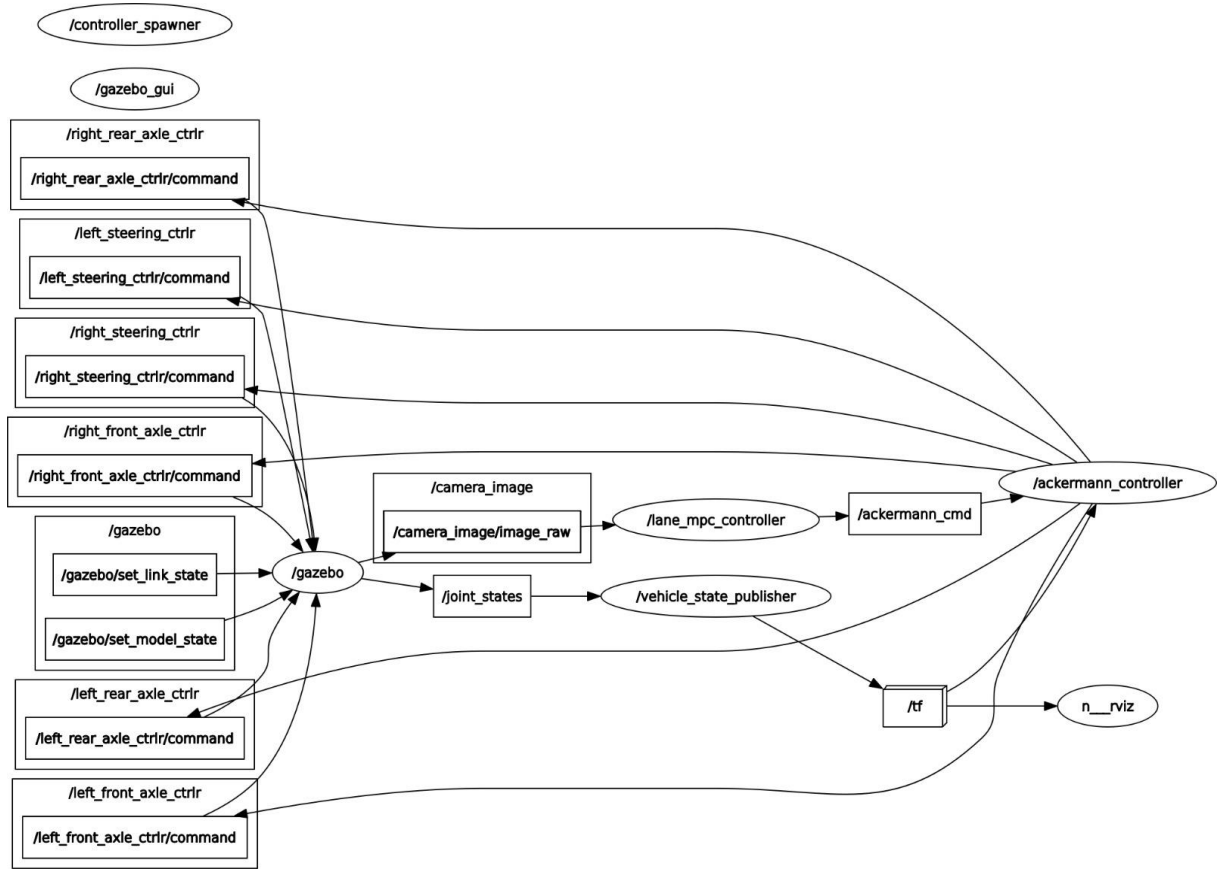

**Figure 19: Ros Structure**

# Chapter 5

## RESULTS & ANALYSIS

During the simulation the following results were obtained. Specially we look at the Yaw rate and RMSE error from the planned path. Several parameters were tweaked to get the final result for both MPC and PID. The PID method was also produced keeping in mind the integral overshoot problem to deal with the nonlinear nature and low controllability of the system.

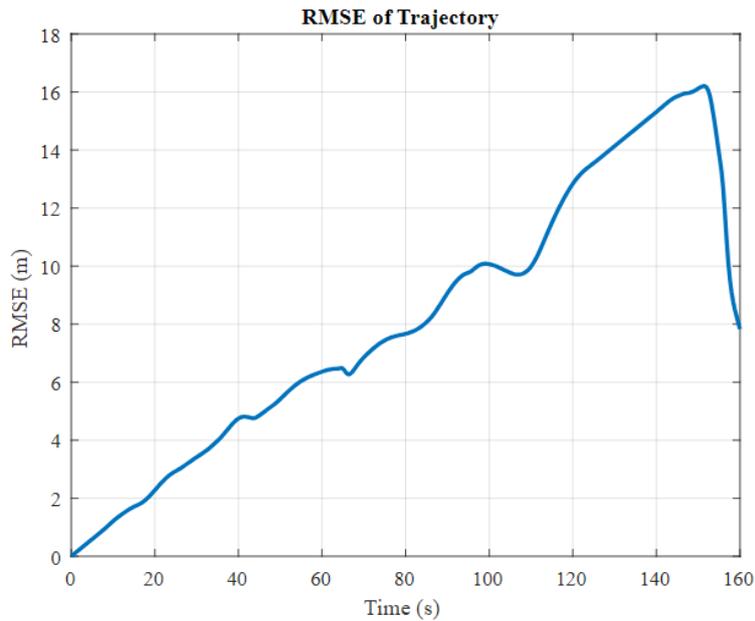

Figure 20: RMSE of the Trajectory

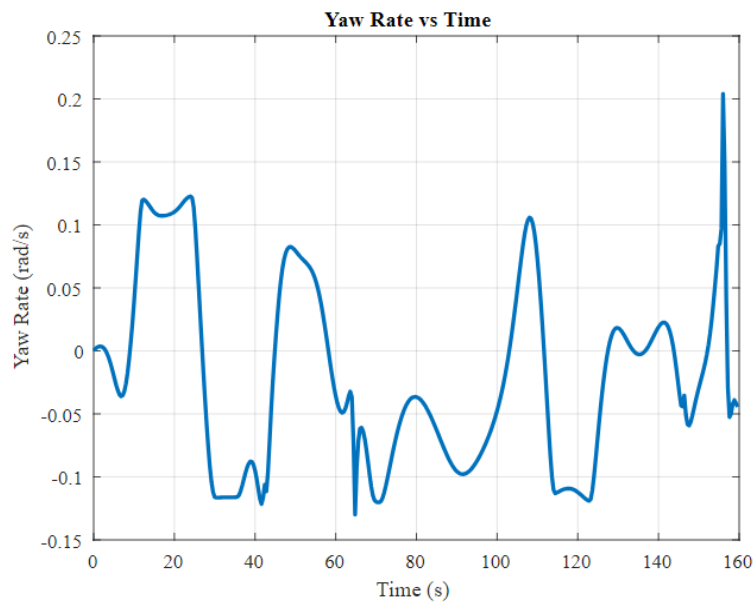

Figure 21: Yaw rate vs time

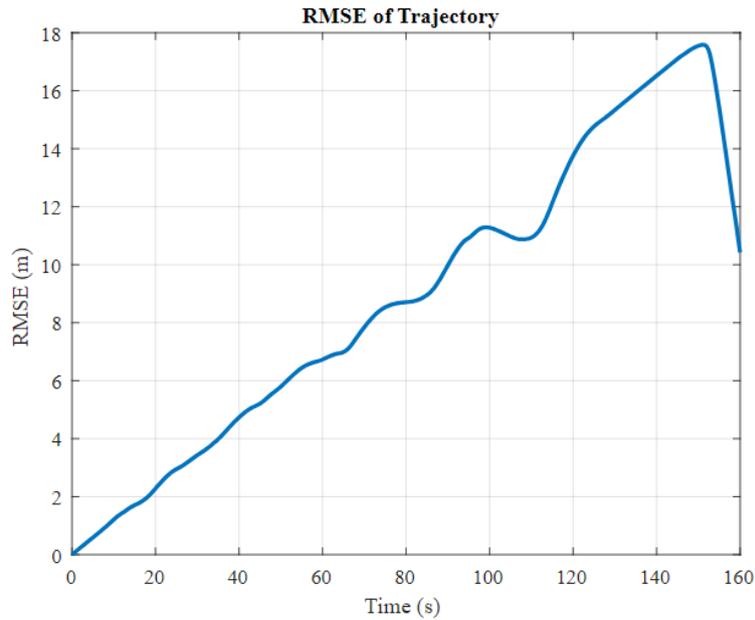

**Figure 22: PID RMSE of the Trajectory**

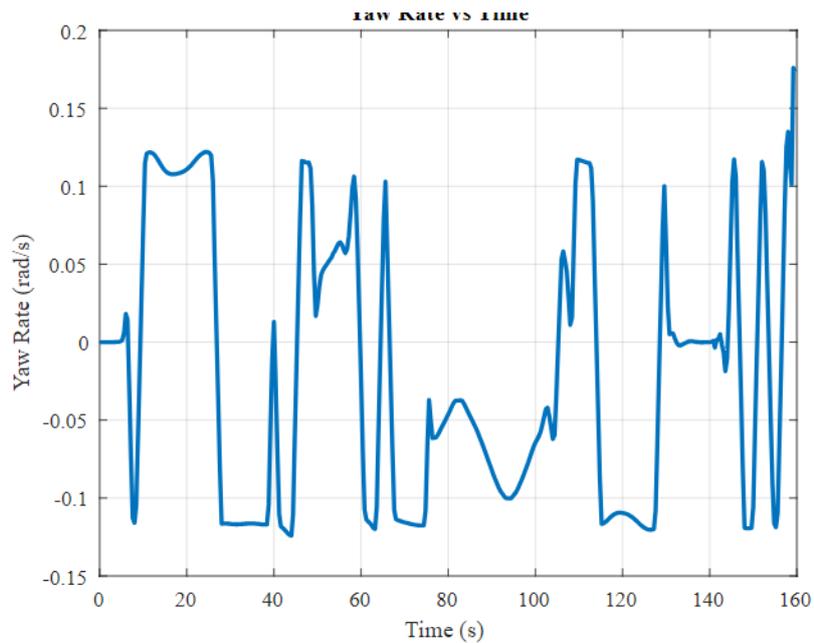

**Figure 23: PID yaw rate vs time**

As seen in the simulation the MPC and PID RMSE errors were similar. PID outperformed the MPC model in the RMSE case. However, let's look at the yaw rate of PID. It is very chaotic and abrupt, whereas MPC provides a smoother yaw rate ensuring this is controllable even in boisterous environments.

So, for a big and non-linear model that is heavy in inertia, MPC shows greater results than PID in simulation.

# Chapter 6

# DEMONSTRATION OF OUTCOME BASED EDUCATION (OBE)

## 6.1 Complex Engineering Problem Addressed:

| Problem | P1 | P2 | P3 | P4 | P5 | P6 | P7 |
|---|---|---|---|---|---|---|---|
| Effectively finding the reference path | * | * |  | * |  |  |  |
| Calculation of the required inputs and system analysis | * |  | * |  |  |  |  |
| Measurement and data analysis and telemetry | * |  |  |  | * |  |  |
| Controllability of the system | * | * |  |  |  |  | * |
| Development of the hardware platform | * |  |  |  |  | * |  |

Effectively Finding the reference path: **P1)** Requires depth knowledge of image segmentation or adjustments techniques that extracts lane boundaries. First, we rotate and undistort the captured image and apply different transformations. Then we take the upper half of the image and extracted lane line pixel positions. **P2)** Here, machine learning algorithms for image segmentation become too computationally heavy for Nvidia jetson and so we looked for simpler methods like sliding window method. **P4)** Often unseen errors can come which results in debugging very tough and lengthy. Like in our case, motor encoders didn't give accurate information about speed and so we shifted to using Pixhawk to get the ground speed.

Calculation of the required inputs and system analysis: **P1)** Depth knowledge regarding the variables to be considered. During MPC testing, the output wasn't converging. Then we have to figure out how to set the constrains to get targeted result. **P2)** Some constrains can lead to better results but does not represent real life scenario. **P3)** Since different inclusion of more components can add different new variables into consideration, so there's no obvious solution. Models get different upon every modification.

Measurement and data analysis and telemetry: **P1)** One has to figure out how a sensor measures. Otherwise, one might end up with offset or shifted data. Like the effect of mismatch height of camera from ground and camera angle to the ground in depth analysis can result in inaccurate depth calculation. Same case goes for camera calibration. **P4)** All the software and products we use are open sources and it follows the standards of scientific community.

Controllability of the system: **P1)** The MPC upon image feedback from camera and current speed, steering angle of autonomous vehicle decides the next speed and velocity. It composes of complex mathematical equations which requires in depth knowledge to get optimized values. **P2)** Often image from camera contains noise, it can end up unsatisfactory output as other variables conflict while looking for optimized values. **P7)** Since controlling the autonomous vehicle is a high-level problem, MPC controls the whole system, it does with the help of feedbacks. It contains small problem which is solved by components by Pixhawk that provides speed, camera that provides real-time images etc.

Development of Hardware platform: **P1)** The hardware platform is a complex combination of different components that serve totally different purposes. Depth knowledge is required to set up proper communication among them so that they can work in harmony. **P6)** Autonomous vehicle has large demand starting from industrial automation to delivering goods to customer doorsteps. Driverless vehicles are the most demanding technology right now.

## 6.2 Course Outcomes (COs) Addressed:

**CO1:** Since working with actual cars for autonomous driving implementation is difficult, expensive, and not within the research capability of common people. A scaled version of the car with maximum similarities can be used for research and development. Platform like Xtenth-car can help us to implement our autonomous driving in a scaler version of a car which will give as responses / outputs similar to an actual sized car. (**P02**)

**CO2:** The functional requirements include a feedback system where the inputs contain accurate information of the speed, angle of steering of the autonomous vehicle as well as camera feedback of what's in the front of the autonomous vehicle. Information is feed to MPC to get optimized speed and angle of steering for better path tracking. The experiments include swift steering movement to travel a path considering passenger comfort as well as maintaining speed to reach the destination quickly. Steering constrains are much emphasized here. (**PO4**)

**CO3:** We are basically creating a testbed with open-source software and products so that the scientific community can replicate it and can be used for further development by the students. Since the overall hardware cost is comparatively less compared to the ones we found from different research papers, it can be a viable alternative. Also, we will make the code and documentation available in GitHub so that others can get support from this. The implementation doesn't have nothing to do with religion since this technology is open for all and can be used alongside with one's own religious views and opinion. (**PO8**)

**CO4:** In modern times, control system algorithms like Model Predictive Control are getting popular due to its robust nature of optimizing. Using state of the art devices that can process images better than others like Nvidia Jetson nano helps to run everything smoothly. MPC can run properly in this. For velocity measuring Pixhawk is being used which is quite popular in making custom drones for its accuracy as well as controlling. A monocular camera is used which is cheap but provides enough resolution for information extraction from images. (**PO5**)

**CO5:** Our capstone project was completed within 1 year. Our work was divided into 3 parts and was distributed among the 3 of us. One did the camera calibration and image processing. Another one implemented the MPC in the nano. Another one figured out the telemetry values from Pixhawk. All of us were going through research papers to get the best performance to cost product.

Version 1 would have costed much less but since it has some serious drawback, we have some upgrade over the processing unit which is Nvidia jetson nano which costs around 350 dollars. For telemetry we used Pixhawk which is 100 dollars. Both of them provides much reliability than the previous alternatives. Overall costing is about 700 dollars for the whole setup. Comparing with the prevailing ones in papers ours is at least 2-4 times cheaper.

For economic decision making, we went for the least expensive product that barely meets our requirements. Then we climbed up depending upon whether the component is suitable or not. It took us some time as we went through several iterations, though we categories them with 2 versions only.

For management, we took some steps like making Gantt chart to figure out our deadlines and progress. We were in constant communication and kept proper documentation on the way to solve problems figured out earlier, faster. **(PO11)**

**CO6 :** People are getting introduced to driverless vehicle. Different companies like Tesla are leading the world on this popular trend. So, people are accepting this. Considering from point of health and safety, this system is still not fully reliable. But still we have many safety features implemented here. Like vehicle stops when the feedback information is out of their desired range. Also since one of our main objectives is passenger comfort, they will feel more comfortable during journey which is good for their health as stress free journey is often appreciated since people often feel sickness during long journey. **(PO6)**

**CO7:** Our proposed system also tries to reduce time on road by proper steering and speed maintaining. So, less fuel will be consumed resulting in less harmful for the environment since consumption of fuel directly or indirectly damages the environment ($CO_2$ increase). Since the proposed system works on every vehicle with any kind of fuel, its long time sustainable. **(PO7)**

**CO9:** Proper communication among the team members helped us to realize the bottleneck of the system. So, we went through different components suitable to our need. We often shared what problems we faced and sometimes one came up with solutions others faced. Apart from our distributed workload, one of us took responsibilities of all hardware manipulation, where other one was looking through programming loopholes. Another one was more focused in research paper findings and how components communicate with each other's. **(PO9)**

**CO10:** Most of our problems that we solved were documented properly so that repetition of the same problem took less time to solve. We communicated with people of different domains who serve different scientific community to get knowledge from their experience about different topics related to our project. **(PO10)**

**CO11:** Every day the technology is getting better. So, there will always be a better solution. So lifelong self-education as well as institutional education is required to keep up with the modern technologies and discovering better solutions. To earn knowledge, meetings with professional people, seminars etc. are much important as they provide insights of what the future might look like. **(PO12)**

## 6.3 Knowledge Profiles (K3 – K8) Addressed

**K3:** An autonomous vehicle requires observation of surroundings which helps it to process and give an output based on the given objective in the system. It can be faster speed, traffic recognition, swift movement etc. So fundamental study of computer vision, control system, power supply, electronics, communication protocol, c and python programming etc. were required.

**K4:** knowledge about embedded systems, circuit designing, programming is required to work.

**K5:** Design should be a close looped system. Depending upon current condition, the future conditions are to be determined. The system should have input of environment and output will be responses to the environment (like motor speed, servo angle of steering)

**K6:** Technology related to control system is the focus here. Control system like MPC has great demand where automation is done. For example, automation in industries for producing goods etc.

**K7:** Our proposed work will be open for all. It will have safety features. It will produce less carbon emission to the environment.

**K8:** People have been working on comparison of different control systems and highlighting the better one. But the tested is expensive. Our purpose is almost similar but we are making a cheaper version. Out improvement is that in the future, we will compare with other control system algorithms to see which one is best in the current scenario and objective.

## 6.4 Attributes of Ranges of Complex Engineering Activities (A1 – A5) Addressed

**A1 (RANGE OF RESOURCES):** We fetched knowledge from different research papers as well as well went through different online resources like Wikipedia, article, project archives, YouTube etc. We have to people who are currently in robotics domain about this to get their opinion and insights. Materials and equipment are mostly funded by our university rover team and rest is funded by the institution authority.

**A2 (Level of Interaction):** the steering angle have much impact on the traversing use of higher steering angle causes jerking and accumulate errors where as in case of low steering angle the vehicle couldn't follow the prediction horizon correctly. Similarly, while preprocessing the image the thresholding value have conflicting output on the result, and thus an optimal value should be chosen.

**A3(Innovation):** Implementation of MPC for autonomous vehicle driving in a compact space is quite novel on its ground because in Bangladesh implementation of this such system is quite impossible but we went through and made it possible. The materials we used are easily available in our country and anyone can replicate it without much guidance.

**A4(Consequences of society and the environment):** The whole system is based on electricity. Since it has no demand of non-renewable energy, its environment friendly. Also, less power consumption results less electricity consumption and thus less $Co_2$ emission.

**A5(Familiarity):** Future improvement can be accomplished by implementing different methods following similar fundamental principles but different approaches for reducing errors and noises from the system.

# Chapter 7